\documentclass[runningheads]{llncs}
\usepackage{graphicx}
\usepackage{amsmath,amssymb,amsfonts}
\usepackage{bm}
\DeclareMathOperator{\sign}{sign}
\usepackage[ruled]{algorithm2e}
\usepackage{multirow}
\usepackage{textcomp}
\usepackage{xcolor}
\usepackage[caption=false]{subfig}

\begin{document}

\title{Boosting Ridge Regression for High Dimensional Data Classification }

%
\author{Jakramate Bootkrajang }
%
%
\institute{
Data Science Research Center, Department of Computer Science,\\ Chiang Mai University, Thailand, 50200 \\
\email{jakramate.b@cmu.ac.th}
}

\maketitle              

\begin{abstract}
Ridge regression is a well established regression estimator which can conveniently be adapted for classification problems. One compelling reason is probably the fact that ridge regression emits a closed-form solution thereby facilitating the training phase. However in the case of high-dimensional problems, the closed-form solution which involves inverting the regularised covariance matrix is rather expensive to compute. The high computational demand of such operation also renders difficulty in constructing ensemble of ridge regressions. In this paper, we consider learning an ensemble of ridge regressors where each regressor is trained in its own randomly projected subspace. Subspace regressors are later combined via adaptive boosting methodology. Experiments based on five high-dimensional classification problems demonstrated the effectiveness of the proposed method in terms of learning time and in some cases improved predictive performance can be observed.
\end{abstract}

\section{Introduction}
Ridge Regression Classifier (RRC) is an adaptation of ridge regression technique widely used in regression problems for classification tasks. 
For binary classification, the adaptation simply involves encoding the response variables of the standard regression formulation with $\pm1$ to represent class label assignments.
The RRC also enjoys the closed-form solution provided by ridge regression method. 
Therefore, learning RRC is usually faster than learning classification models which rely on iterative optimisation. 
Although, RRC works quite well in low-dimensional problems, the method is somehow less favourable in today's increasingly common high-dimensional data such as classification of microarray \cite{wang2008approaches}, social network text  \cite{ghaddar2018high} or hyperspectral images \cite{gao2018spectral} to name a few. 
The major reason why RRC is deemed inappropriate for high-dimensional data could be the fact that its closed-form solution involves matrix inversion which unfortunately does not scale well with data dimensionality. 

High computational demand of learning RRC via the closed-form solution also poses difficulty in combining multiple RRCs for high-dimensional data classification. 
For example, previous studies on boosting ridge regression only consider problems with the number of features not exceeding 200 \cite{tutz07boosting}. This drawback greatly limits the applicability of RRC as well as its ensemble to various interesting high-dimensional classification tasks. 
One way to mitigate the problem is to perform dimensionality reduction prior to learning the classifier. Nonetheless, dimensionality reduction methods, such as Principal Component Analysis (PCA), also inevitably suffers from the curse of dimensionality. In addition, one should be aware that the errors occurred during such pre-processing steps might unnecessarily be accumulated. 
Inspired by the work in \cite{kaban2014new} where random projection technique was successfully employed in solving high-dimensional ordinary least square problem with theoretical guarantee, in this work we study a strategy for efficiently constructing an ensemble of RRCs for high-dimensional data. Essentially, the strategy involves obtaining a classifier in random subspace induced by random projection matrix. Since the dimension of subspace can be much smaller than that of the original space, the learning can be accelerated. Subspace classifiers are then projected back to the original feature space and get combined. In contrast to existing work which combines the subspace classifiers by means of averaging, we will instead explore an alternative of combining the subspace classifiers via adaptive boosting methodology \cite{schapire2013explaining}. 

The rest of the paper is organised as follows. Section \ref{sec:method} presents the background and the algorithm for boosting subspace ridge regressions. The empirical evaluation and the discussion of the experimental results are presented in Section \ref{sec:exper}. Finally, Section \ref{sec:conclusion} concludes the study and outlines the future work.

\section{Background and Method}
\label{sec:method}

\subsection{Ridge regression classifier}
Let $\bold{X} \in \mathbb{R}^{N\times d}$ be a row-major data matrix of $N$ input instances in $d$-dimensional feature space, and $\bold{X}^T$ its transpose. Let $\bold{y} \in \mathbb{R}^N$ be a column vector of response variables. Given $\bold{X}$ and $\bold{y}$, the objective of ordinary least square problem is to find a column vector $\boldsymbol{\beta} \in \mathbb{R}^d$ (i.e., model parameters) that minimises the discrepancy between $\bold{X}\boldsymbol{\beta}$ and $\bold{y}$. Mathematically, we aim for $\boldsymbol{\beta}$ which minimises $\frac{1}{2}||\bold{X}\boldsymbol{\beta} - \bold{y}||^2_2$. To adapt ridge regression for binary classification, we let $y_i$, an element of $\bold{y}$, takes value from $\{-1,1\}$, while the objective function remains unchanged.
The optimal $\boldsymbol{\beta}$ are found by first taking derivative of the objective function w.r.t. $\boldsymbol\beta$.
\begin{align}
\frac{\partial \frac{1}{2}||\bold{X}\boldsymbol{\beta} - \bold{y}||^2_2}{\partial \boldsymbol\beta} &= (\bold{X}\boldsymbol{\beta} -\bold{y})\bold{X}
\end{align}
Equating the above to zero and solving for $\boldsymbol\beta$ yields the closed-form solution for the model parameters.
\begin{align}
\boldsymbol{\hat\beta} &= (\bold{X}^T\bold{X})^{-1}  \bold{X}^T\bold{y}
\label{eq:lsqr_obj}
\end{align}
For high-dimensional and small sample size data such as microarray gene expression data, it is very likely that the covariance matrix $\bold{X}^T\bold{X}$ will be ill-conditioned, and thus its inverse cannot be accurately estimated, if not incomputable at all. Ridge regression imposes an independence assumption between features by adding a `ridge' on the diagonal elements of the covariance matrix so that Eq.(\ref{eq:lsqr_obj}) becomes
\begin{align}
\boldsymbol{\hat{\beta}} &= (\bold{X}^T\bold{X} + \lambda \bold{I})^{-1} \bold{X}^T\bold{y},
\end{align}
where $\lambda$ is a regularisation parameter and $\bold{I}$ is a $d{\times}d$ identity matrix. The above solution is more stable but still relies on costly matrix inversion of which its complexity is $O(d^3)$. 

\subsection{Boosting subspace RRCs}
We have witnessed that learning a single RRC is already quite time consuming. It is therefore inefficient to form an ensemble of RRCs especially in the case of high-dimensional data. This greatly limits the applicability of the RRC model in real world. 
To alleviate the problem, we consider learning an RRC in a lower $m$ dimensional subspace induced by random projection matrix. Subspaces classifiers are then additively combined via Adaptive Boosting (AdaBoost) methodology. 

Let us consider a binary classification where $y_i \in \{-1,1\}$. Let $L(\boldsymbol\alpha)$ represents the exponential loss function of AdaBoost
\begin{align}
L(\boldsymbol\alpha) &= \sum_{i=1}^Nw_i\exp\Big(-y_i\sum_{k=1}^K \alpha_k h_k(\bold{x}_i)\Big).
\label{eq:obj}
\end{align}
Here, $\alpha_k$ represents the weight for the outputs contributed by $h_k(\bold{x}) = \sign(\bold{x}^T\boldsymbol\beta_k)$, the $k$-th ridge regression classifier. And $w_i$ is the weight of the $i$-th input vector.  
At round $k$ of boosting, minimising Eq.(\ref{eq:obj}) is equivalent to finding $h_k()$ that minimises the weighted sum of misclassification errors, $\epsilon_k = \sum_{i=1}^N w_i \mathbf{1}(h_k(\bold{x}_i)\neq y_i)$. Now instead of solving for $\boldsymbol\beta_k$ in the original feature space, we shall approximate $\boldsymbol\beta_k$ by the average of multiple subspace RRCs trained in the randomly projected subspaces. Denoting subspace RRC by $\boldsymbol{b}$, it can be shown that $\boldsymbol{b}$ which minimises the weighted error is given by
\begin{align}
\boldsymbol{\hat{b}} = (\bold{Z}^T\bold{W}\bold{Z} - \lambda \bold{I})^{-1} \bold{Z}^T\bold{W}\bold{y}
\label{eq:wls_sol}
\end{align}
where $\bold{Z} := \bold{X}\bold{R}$ is the projected data matrix, and $\bold{R}$ is an $d{\times}m$ random matrix with elements drawn i.i.d from a normal distribution, $\mathcal{N}(0,1/d)$ \cite{vempala2005random,kaban2014new}. The matrix $\bold{W}$ is a diagonal matrix of size $N{\times}N$ with $w_i$ as its diagonal elements. Reader may recall that Eq.(\ref{eq:wls_sol}) is in fact the solution to the weighted least square problem. To this end, we recover RRC in the original space by averaging, $\boldsymbol{\hat{\beta}_k} = \sum_{p=1}^P \boldsymbol{b}_p/P$.
The contribution of $h_k()$ towards the prediction of the ensemble can be calculated with
$ \alpha_k = \frac{1}{2}\ln\Big(\frac{1-\epsilon_k}{\epsilon_k}\Big) $.
Finally, we direct the next RRC towards misclassified examples by adjusting the weight of the input data
\begin{align}
w^{k+1}_i = w^k_i \exp{(-y_i\alpha_k h_k(\bold{x}_i))}
\end{align}
To predict the label of an unseen example $\bold{x}_q$, we decide
$ \hat{y}_q = \sign\Big(\sum_{k=1}^K \alpha_k h_k(\bold{x}_q)\Big)$.
Algorithm \ref{alg:rpboost} summarises the steps for boosting subspace RRC. 
\begin{algorithm}
\SetAlgoNoLine
\caption{Boosting subspace ridge regression classifiers.}
\footnotesize
\DontPrintSemicolon
\KwIn {Data matrix $\bold{X}^{N\times d}$, label vector $\bold{y}^{N\times 1}$, boosting rounds $K$, random projection rounds $P$ and target dimension $m$.}
\Indp Initialise the weight of data instance to $w_i = 1/N$ \;
 Initialise the weight matrix $\bold{W}^{N \times N} = diag(w_{i=1:N})$ \;
\For{$k\gets1$ \KwTo $K$ }{
	\For{$p\gets1$ \KwTo $P$ }{
		Create a random matrix $\bold{R}^{d\times m}$ where $r_{ij} \sim N(0,1/d) $ \;
		Perform random projection on data, $\bold{Z}^{N\times m} = \bold{X}\bold{R}$ \;
		Estimate subspace RRC by $\boldsymbol{\hat{b}}_p = (\bold{Z}^T\bold{W}\bold{Z} - \lambda \bold{I})^{-1} \bold{Z}^T\bold{W}\bold{y}$\;
	}
	Recover RRC by $\boldsymbol{\hat{\beta}_k} = \sum_{p=1}^P \boldsymbol{b}_p/P$\;
	Compute weighted misclassification error $\epsilon_k = \sum_{i=1}^N w_i \mathbf{1}(h_k(\bold{x}_i) \neq y_i)$\;
	Compute contribution $\alpha_k = \frac{1}{2}\ln\Big(\frac{1-\epsilon_k}{\epsilon_k}\Big)$\;
	Update the data weights $w^{k+1}_i = w^k_i \exp{(-y_i\alpha_k h_k(\bold{x}_i))}$ and $\bold{W}$\;
    }
\Indm
\KwOut{$\boldsymbol\alpha, \boldsymbol\beta_{k=1:K}$}
\label{alg:rpboost}
\end{algorithm}

\section{Empirical evaluations}
\label{sec:exper}

In this section, we present the empirical evaluations of the proposed boosting strategy for high-dimensional classification problems.
Specifically, we compared the proposed strategy with the traditional boosting approach, i.e., RRCs were directly trained in the original feature space, and with single RRC. We then measured the running times and the generalisation errors. 
The results should shed light on the tradeoff between speeding up the learning and the predictive performance of the model as well as on how efficient the proposed strategy is compared to learning a single RRC in high dimensional data settings.  

For that purpose, we employed 5 high-dimensional datasets from various application domains namely image classification (\textit{websearch}\cite{Bootkrajang12Label}), text classification (\textit{ads}), and gene expression classification (\textit{colon}\cite{Alon99Broad}, \textit{breast-cancer}\cite{West01Predicting} and \textit{leukaemia}\cite{Golub99molecular}). The characteristics of the datasets including the number of training instances and data dimensionality are summarised in Table \ref{tab:data}. For each run, we randomly split the data into training set and test set using 80/20 train/test splitting ratio. We then recorded training time and classification accuracies. We repeated 20 independent runs according to the above protocol to get reliable statistics. 
We empirically set $\lambda = 0.3$, $m = 3$ and $P = 3$. We allowed 300 of boosting rounds. All of the experiments were conducted on a machine equipped with Intel(R) Core(TM) i5-4570 CPU @ 3.20GHz and 8GB of RAM.

\begin{table}[t!]
\centering
\caption{The characteristics of the datasets employed in this study. Add class-based statistics}
\begin{tabular}{|c | c | c |c | c | c |}
\hline
 & \textit{websearch} &  \textit{ads} & \textit{colon} & \textit{breast} & \textit{leukaemia} \\
\hline
\# of instances & 1030 & 3279 & 62 & 49 & 72 \\
\# of features & 1318 & 1558 & 2000 & 7129 & 7129 \\
\hline
\end{tabular}
\label{tab:data}
\end{table}

\subsection{Results and discussion}

Table \ref{tab:rpboost_vs_boost} presents the averaged learning times, mean generalisation errors and standard errors of the proposed boosting strategy (rpBoost), traditional RRC boosting (RRC-Boost) and single RRC.
\begin{table}[h!]
\centering
\caption{Learning times and generalisation errors of RRC and rpBoost algorithms.}
\begin{tabular}{|c | c |c | c | c| c | c |c |}
\hline
\multirow{2}{*}{Dataset} & \multicolumn{3}{c|}{learning time (seconds.)} &  \multicolumn{3}{c|}{generalisation error} \\ \cline{2-7}
& RRC & RRC-Boost & rpBoost & RRC & RRC-Boost & rpBoost \\ \hline
\textit{websearch} & 0.07$\pm$0.00 & 34.56$\pm$0.19 & 2.44$\pm$0.06 & 0.27$\pm$0.04 & 0.17$\pm$0.03 & 0.16$\pm$0.02\\
\textit{ads} & 0.23$\pm$0.01 & 171.08$\pm$4.47 & 17.68$\pm$0.08 & 0.03$\pm$0.01 & 0.04$\pm$0.01 & 0.06$\pm$0.01\\
\textit{colon} & 0.09$\pm$0.00 & 50.88$\pm$0.22 & 0.28$\pm$0.02 & 0.10$\pm$0.10 & 0.10$\pm$0.11 & 0.09$\pm$0.08\\
\textit{breast} & 1.96$\pm$0.01 & 714.23$\pm$1.00 & 0.53$\pm$0.00 & 0.15$\pm$0.09 & 0.16$\pm$0.09 & 0.13$\pm$0.09 \\
\textit{leukaemia} & 1.97$\pm$0.01 & 720.44$\pm$0.61 & 0.63$\pm$0.06 & 0.06$\pm$0.05 & 0.06$\pm$0.05 & 0.08$\pm$0.08 \\ \hline
\end{tabular}
\label{tab:rpboost_vs_boost}
\end{table}
Let us first examine the time needed to train the models in the chosen datasets. It is evident that rpBoost improved over traditional RRC-Boost in all of the cases tested. For example, RRC-Boost with 300 rounds would take around 50 seconds to train on \textit{colon} dataset whereas rpBoost took less than half a second. It is interesting that training RRCs using the proposed rpBoost in extremely high-dimensional problems was also more efficient than training a single RRC e.g., as in \textit{breast} and \textit{leukaemia} where the dimensionality of the data exceeds 7000. 
Despite working on the approximated version of input data, we noticed that rpBoost were still comparable to the traditional approaches in terms of predictive accuracy. Surprisingly, we observed slight improvements in generalisation errors in 3 out of 5 cases i.e., \textit{websearch}, \textit{colon} and \textit{breast}. In the rest of the cases, the performances of rpBoost were not much lagged behind. The results hint that the proposed boosting strategy is a promising method for high-dimensional data classification. 

To further investigate the effectiveness of rpBoost, we compared it to AdaBoost with decision stump (Stump) and a random projection-based technique akin to that presented in \cite{kaban2014new} but with RRC as base classifier instead of ordinary least square. The method essentially combines subspace classifiers using averaging, i.e., $\boldsymbol{\hat{\beta}} = \sum_{l=1}^L \boldsymbol{b}_l/L$ where $L$ is set to 300 i.e., equivalent to the boosting rounds. We will refer to this method as rpRRC. 
\begin{table}[h!]
\centering
\caption{Performance comparison with boosted decision stumps and rpRRC.}
\begin{tabular}{|c | c |c | c | c| c | c |c |}
\hline
\multirow{2}{*}{Dataset} & \multicolumn{3}{c|}{learning time (seconds.)} &  \multicolumn{3}{c|}{generalisation error} \\ \cline{2-7}
& rpRRC & Stump & rpBoost & rpRRC & Stump & rpBoost \\ \hline
\textit{websearch} & 0.25$\pm$0.04 & 30.88$\pm$0.12 & 2.44$\pm$0.06 & 0.14$\pm$0.02 & 0.13$\pm$0.03 & 0.16$\pm$0.02\\
\textit{ads} & 0.95$\pm$0.15 & 81.97$\pm$0.66 & 17.68$\pm$0.08 & 0.12$\pm$0.01 & 0.03$\pm$0.01 & 0.06$\pm$0.01\\
\textit{colon} & 0.14$\pm$0.12 & 7.24$\pm$0.16 & 0.28$\pm$0.02 & 0.13$\pm$0.13 & 0.08$\pm$0.05 & 0.09$\pm$0.08\\
\textit{breast} & 0.22$\pm$0.04 & 23.97$\pm$0.17 & 0.53$\pm$0.00 & 0.15$\pm$0.11 & 0.16$\pm$0.11 & 0.13$\pm$0.09 \\
\textit{leukaemia} & 0.18$\pm$0.02 & 27.35$\pm$0.09 & 0.63$\pm$0.06 & 0.19$\pm$0.10 & 0.07$\pm$0.06 & 0.08$\pm$0.08 \\  \hline
\end{tabular}
\label{tab:rpboost_vs_stump}
\end{table}
It can be learned from the results in Table \ref{tab:rpboost_vs_stump} that rpRRC is clearly the fastest algorithm because there were no instance weights and learner contributions updating as in rpBoost. This, however, slightly sacrifices the predictive performance. Boosted decision stumps seemed to give the best generalisation errors but is at the same time the least efficient method. The proposed rpBoost tends to balance both learning time and generalisation errors. Nonetheless, we believe that the extra headroom in learning time is beneficial for rpBoost to reach or even surpass the performance of existing methods when more boosting rounds are available while still being faster to train. 
Lastly, it is worth noting the interplay between $m$ the subspace dimension and $P$ the number of subspace RRCs. Generally, decreasing $m$ reduces the training time while increasing the variance of estimators due to the data being more distorted. 
The variance can be controlled by increasing $P$ which, in turn, increases the training time. To the best of our knowledge, there is currently no theoretical guidance for setting these two hyperparameters and practitioners therefore resort to choosing $m$ and $P$ empirically. 

\section{Conclusions}
\label{sec:conclusion}
We presented an approach for speeding up the boosting of ridge regression classifiers in high-dimensional space through the use of random projection technique. 
Each RRC member of the ensemble was approximated by multiple subspace RRCs. The RRC members are then combined via adaptive boosting methodology.
The experimental results demonstrated that the random projection can speed up the learning while also maintaining generalisation performance compared to standard ridge regression boosting and single ridge regression classifier. Theoretical analysis of rpBoost is surely worth exploring in the future. 

\bibliographystyle{splncs04}
\bibliography{esann2020}

\end{document}